\documentclass[journal]{IEEEtran}
\usepackage{cite}
\usepackage{graphicx}
\usepackage{textcomp}
\usepackage{caption}
\usepackage{subcaption}
\usepackage{xspace}
\usepackage[cmex10]{amsmath}
\usepackage{amsthm}
\usepackage{cite}
\usepackage{algorithm}
\usepackage{algpseudocode}
\usepackage{array}
\usepackage{url}
\usepackage{dblfloatfix}
\usepackage{multirow}
\usepackage{color}
\usepackage{epstopdf}
\usepackage{endnotes}
\usepackage{framed}
\usepackage{flushend}
\usepackage{cool} 
\usepackage{enumerate} 
\usepackage{gensymb}
\usepackage{hyperref}

\usepackage{hyperref}       
\usepackage{url}            
\usepackage{multirow} 
\usepackage{booktabs}       
\usepackage{amsfonts}       
\usepackage{nicefrac}       
\usepackage{microtype}      
\usepackage{varwidth}
\usepackage[switch]{lineno}

\newcommand{\DemLearn}{\ensuremath{\textsf{DemLearn}}\xspace}

\begin{document}

\title{Self-organizing Democratized Learning: Towards Large-scale Distributed Learning Systems}
	\author{~Minh~N.~H.~Nguyen,\IEEEmembership{~Member,~IEEE},~Shashi~Raj~Pandey,\IEEEmembership{~Member,~IEEE,}~Tri~Nguyen~Dang,\\~Eui-Nam~Huh,\IEEEmembership{~Member,~IEEE},~Nguyen~H.~Tran,\IEEEmembership{~Senior~Member,~IEEE},\\
		~Walid~Saad,\IEEEmembership{~Fellow,~IEEE},~and~Choong~Seon~Hong,\IEEEmembership{~Senior~Member,~IEEE}.
	
	\IEEEcompsocitemizethanks{	
			This work was supported by the National Research Foundation of Korea(NRF) grant funded by the Korea government(MSIT) (No. No. 2020R1A4A1018607) and by Institute of Information \& Communications Technology Planning \& Evaluation (IITP) grant funded by the Korea government(MSIT) (No.2019-0-01287, Evolvable Deep Learning Model Generation Platform for Edge Computing), and the Institute of Information and Communications Technology Planning and Evaluation (IITP) Grant funded by the Korea Government (MSIT) (Artificial Intelligence Innovation Hub) under Grant 2021-0-02068. (\textit{Corresponding authors: Choong Seon Hong, Huu Nhat Minh Nguyen, and Eui-Nam Huh.})
			
			\IEEEcompsocthanksitem M.~N.~H.~Nguyen is with The University of Danang - Vietnam-Korea University of Information and Communication Technology, Vietnam and also the Department of Computer Science and Engineering, Kyung Hee University, South Korea. (email: nhnminh@vku.udn.vn).
			\IEEEcompsocthanksitem T.~D.~Nguyen,~E.~N.~Huh,~and~C.~S.~Hong are with the Department of Computer Science and Engineering, Kyung Hee University, Yongin-si 17104, South Korea. (email: \{trind, shashiraj, johnhuh, cshong\}@khu.ac.kr).
			\IEEEcompsocthanksitem S.~R.~Pandey is with the Connectivity Section, Department of Electronic Systems, Aalborg University, Denmark and also the Department of Computer Science and Engineering, Kyung Hee University, South Korea. . (email: srp@es.aau.dk).
			\IEEEcompsocthanksitem N. H. Tran is with School of Computer Science, The University of Sydney,
Sydney, NSW 2006, Australia (email: nguyen.tran@sydney.edu.au).
			\IEEEcompsocthanksitem W.~Saad is with the Wireless@VT, Bradley Department of Electrical and Computer Engineering, Virginia Tech, Blacksburg, VA, 24060, USA and also with the Department of Computer Science and Engineering, Kyung Hee	University, Yongin 446-701, South Korea. (email: walids@vt.edu).
		}
	}

\markboth{Preprint Version}{}%

	\maketitle

\begin{abstract}
   Emerging cross-device artificial intelligence (AI) applications require a transition from conventional centralized learning systems towards large-scale distributed AI systems that can collaboratively perform complex learning tasks. In this regard, \emph{democratized learning} (Dem-AI) lays out a holistic philosophy with underlying principles for building large-scale distributed and democratized machine learning systems. The outlined principles are meant to study a generalization in distributed learning systems that goes beyond existing mechanisms such as federated learning. 
  Moreover, such learning systems rely on hierarchical self-organization of well-connected distributed learning agents who have limited and highly personalized data and can evolve and regulate themselves based on the underlying duality of \emph{specialized} and \emph{generalized processes}.
   Inspired by Dem-AI philosophy, a novel distributed learning approach is proposed in this paper. The approach consists of a self-organizing hierarchical structuring mechanism based on agglomerative clustering, hierarchical generalization, and corresponding learning mechanism. Subsequently, hierarchical generalized learning problems in recursive forms are formulated and shown to be approximately solved using the solutions of distributed personalized learning problems and hierarchical update mechanisms. To that end, a distributed learning algorithm, namely \DemLearn is proposed. Extensive experiments on benchmark MNIST, Fashion-MNIST, FE-MNIST, and CIFAR-10 datasets show that the proposed algorithm demonstrates better results in the generalization performance of learning models in agents compared to the conventional FL algorithms. The detailed analysis provides useful observations to further handle both the generalization and specialization performance of the learning models in Dem-AI systems.
\end{abstract}


\IEEEpeerreviewmaketitle
\vspace{-0.1pt}

\begin{figure*}[t]
	\centering
	\includegraphics[width=\linewidth]{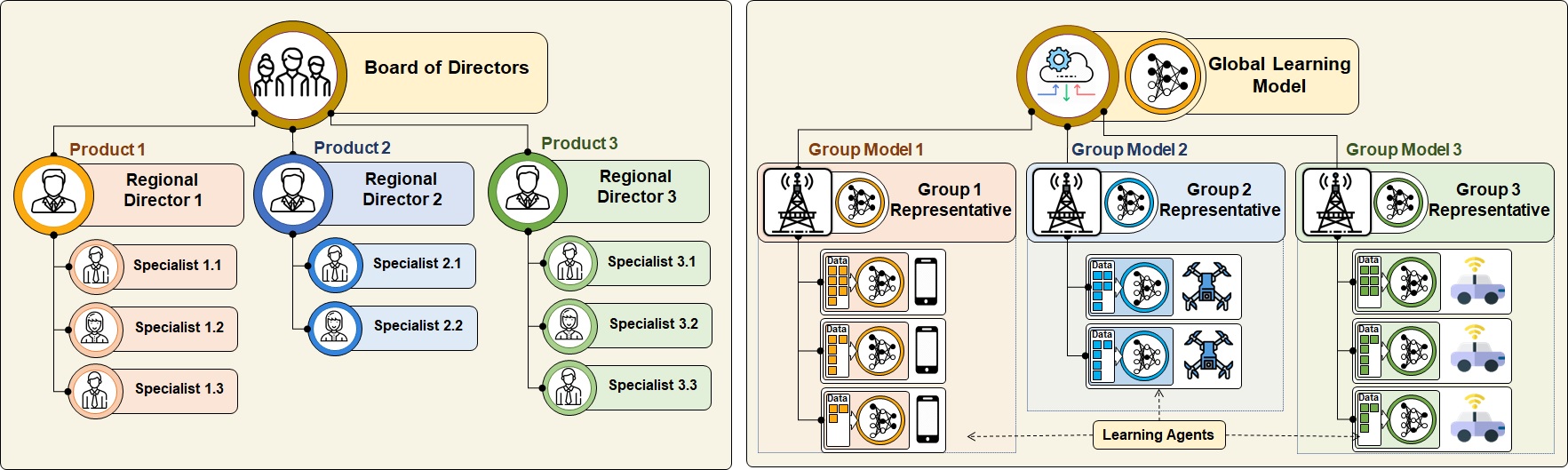}
	\caption{Analogy of a hierarchical distributed learning system.}
	\label{F:Motivation}
\end{figure*}

\section{Introduction}
Nowadays, AI has grown to be successful in solving complex real-life problems such as decision support in healthcare systems, advanced control in automation systems, robotics, and telecommunications. Numerous existing mobile applications incorporate AI modules that leverage user's data for personalized services such as Gboard mobile keyboard on Android, QuickType keyboard, and the vocal classifier for Siri on iOS \cite{FL_advances}. 
By exploiting the unique features and personalized characteristics of users, these applications not only improve the personal experience of the users but also helps to better control over their devices. Moreover, the rising concern of data privacy in existing machine learning frameworks fueled a growing interest in developing distributed machine learning paradigms such as federated learning frameworks (FL) \cite{FL_advances,mcmahan2016communication,mocha2017,tran2019federated,nguyen2020toward,pandey2020crowdsourcing,FL_TON2019,chen2019joint,fallah2020personalized,deng2020adaptiveFL,federatedSurvey2020}. FL was first introduced in \cite{mcmahan2016communication}, where the learning agents coordinate via a central server to train a global learning model in a distributed manner. These agents receive the global learning model from the central server and perform local learning based on their available datasets. Then, they send back the updated learning models to the server for updating the global model via an aggregation operation without revealing the private training data to the others.

 In practice, the private dataset collected at each agent is unbalanced, highly personalized for some applications such as handwriting and voice recognition, and exhibit non-i.i.d (non-independent and non-identically distributed) characteristics. Therefore, the iterative process of updating the global model improves the generalization of the model, but also hurts the personalized performance at the agents \cite{FL_advances}. Hence, existing FL algorithms cannot efficiently handle the underlying cohesive relation between generalization and personalization (or specialization) abilities of the trained learning model \cite{FL_advances}. 
To the best of our knowledge, the work in \cite{fallah2020personalized} was the first attempt to study and improve the personalized performance of FL using a personalized federated averaging (Per-FedAvg) algorithm based on a meta-learning framework (MLF). Furthermore, in a recent work \cite{deng2020adaptiveFL}, the authors propose an adaptive personalized FL framework where a mixture of the local and global model was adopted to reduce the generalization error. However, similar to \cite{fallah2020personalized}, the cohesive relation between generalization and personalization was not adequately analyzed. Recently, \cite{canh2020personalized} proposed pFedMe algorithm by studying a bi-level learning optimization problem such as global problem and personalized problems.

To better analyze the personalized and generalized learning performance for learning models in FL framework, the Dem-AI philosophy, discussed in \cite{Dem-AI} introduces a holistic approach and general guidelines to develop distributed and democratized learning systems. The approach refers to observations about the generalization and specialization capabilities of biological intelligence, and the hierarchical structure of society and swarm intelligence in large-scale distributed learning systems. Fig.~\ref{F:Motivation} illustrates the analogy of the Dem-AI system and the hierarchical structure in an organization. The specialists from different domain knowledge are grouped into teams to perform common products or targets. These groups in an organization need to collaborate towards the common goals under the supervision of a board of directors. Similarly, learning agents in different groups perform the collaborative learning for group models. The outputs of these groups in a Dem-AI system are the specialized learning models that are created by group members. 
In this paper, inspired by Dem-AI guidelines, we develop a novel distributed learning framework that can directly extend the conventional FL scheme for collectively solving a common learning task at learning agents. Different from existing FL algorithms for building a single generalized model (a.k.a global model), we maintain self-organizing hierarchical group models. 
Accordingly, we adopt the agglomerative hierarchical clustering \cite{hierarchical-clustering-review} and periodically update the hierarchical structure based on the similarity in the learning characteristic of users. In particular, we propose the hierarchical generalization and learning problems for each generalized level in a recursive form. To solve the complex formulated problem due to its recursive structure, we develop a distributed learning algorithm, \DemLearn. The proposed algorithm uses the bottom-up scheme to iteratively performs the local learning by solving personalized learning problems and hierarchical update the generalized models for groups at higher levels. With extensive experiments, we validate both specialization and generalization performance of all learning models on benchmark MNIST, Fashion-MNIST, Federated Extended MNIST (FE-MNIST), and CIFAR-10 datasets.

    To that end, we discuss the preliminaries of democratized learning in Section \ref{S:DemAI Preliminary}. Based on the Dem-AI guidelines, we formulate hierarchical generalized, personalized learning problems, and propose a novel distributed learning algorithm in section \ref{S:Learning Design}. We validate the efficacy of our proposed algorithm for both specialization and generalization performance of the client, groups, and global models compared to the conventional FL algorithms in Section \ref{S:Performance Evaluation}. Finally, Section \ref{S:Conclusion} concludes the paper.   

 \begin{figure*}[t]
	\centering
	\includegraphics[width=\linewidth]{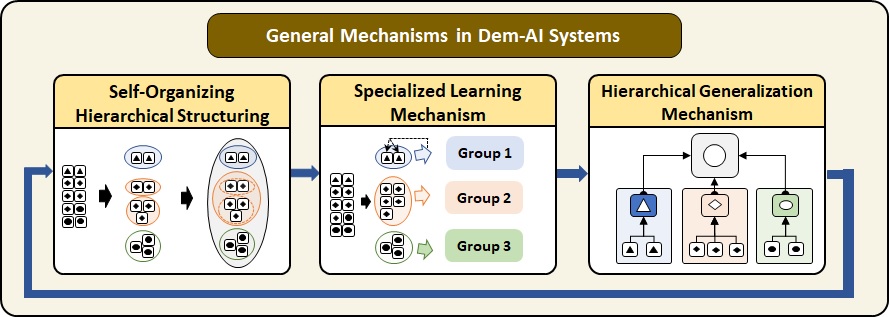}
	\caption{Three general mechanisms for designing Dem-AI systems.}
	\label{F:Dem_AI_Principle}
\end{figure*}

\section{Democratized Learning: Preliminaries} \label{S:DemAI Preliminary}
 Different from FL, the Dem-AI framework \cite{Dem-AI} introduces a self-organizing hierarchical structure for solving common single/multiple complex learning tasks by mediating contributions from a large number of learning agents in collaborative learning. Moreover, it unlocks the following features of \emph{democracy} in the future distributed learning systems.  According to the differences in their characteristics, learning agents form appropriate groups that can be specialized for similar agents to deal with the learning tasks. These specialized groups are self-organized in a hierarchical structure and collectively construct the shared generalized learning knowledge to improve their learning performance by reducing individual biases due to the unbalanced, highly personalized local data. In particular, the learning system allows new group members to: a) speed up their learning process with the existing group knowledge, and b) incorporate their new learning knowledge in expanding the generalization capability of the whole group. In Dem-AI systems, learning agents are free to join any of the appropriate groups and exhibit equal power in the construction of their groups' generalized learning model. Here, the power of each group can be represented by the number of its members which varies over the training time. We introduce a brief summary of Dem-AI concepts and principles \cite{Dem-AI} in the following discussion.

	\textbf{Definition and goal:} Democratized Learning (\textit{Dem-AI} in short) studies a dual (coupled and working together) specialized-generalized processes in a self-organizing hierarchical structure of large-scale distributed learning systems. The specialized and generalized processes must operate jointly towards an \emph{ultimate learning goal} identified as performing collective learning from biased learning agents, who are committed to learning from their own data using their limited learning capabilities. As such, the ultimate learning goal of the Dem-AI system is to establish a mechanism for collectively solving common (single or multiple) complex learning tasks from a large number of learning agents. 
	

	\textbf{Specialized Process:} This process is used to leverage \textit{specialized learning} capabilities at the learning agents and specialized groups by exploiting their collected data. 
	By incorporating the generalized knowledge of higher level groups created by the generalization mechanism, the learning agents can update their model parameters so as to reduce biases in their \emph{personalized learning}. Thus, the personalized learning objective has two goals: 1) To perform \emph{specialized learning}, and 2) \emph{to reuse the available hierarchical generalized knowledge}.

    \textbf{Generalized Process:} 
    The generalization mechanism encourages group members to share knowledge when performing learning tasks with similar characteristics and construct hierarchical levels of generalized knowledge. The hierarchical generalized knowledge helps the Dem-AI system maintain the generalization ability for reducing biases of learning agents and efficiently dealing with environment changes or performing new learning tasks.

	\textbf{Self-organizing Hierarchical Structure:}
	 The hierarchical structure of specialized groups and the relevant generalized knowledge are constructed and regulated following a self-organization principle based on the similarity of learning agents. In particular, this principle governs the union of small groups to form a bigger group that eventually enhances the generalization capabilities of all members. Thus, specialized groups at higher levels in the hierarchical structure have more members and can construct more generalized (less biased) knowledge faster adaptation to new environments in \cite{mengistu2016evolutionaryhierarchy}. 

	\textbf{Transition in the dual specialized-generalized process:}
     The specialized process becomes increasingly important compared to the generalized process during the training time. As a result, the learning system evolves to gain specialization capabilities from the learned tasks but also loses the capabilities to deal with environmental changes such as new learning agents, and new learning tasks. Meanwhile, the hierarchical structure of the Dem-AI system is self-organized and evolved from a high level of plasticity to a high level of stability, i.e., from unstable specialized groups to well-organized specialized groups. 
    The transition of the Dem-AI learning system is illustrated in Fig.~ \ref{F:Dem_AI_Principle} with three iterative sub-mechanisms such as \textit{generalization}, \textit{specialized learning} and \textit{hierarchical structuring mechanism}. Accordingly, the transition of the dual specialized-generalized process represents the steps in a typical democratized learning framework \cite{Dem-AI}.  In that transition, the learning agents are grouped according to the similarities of their learning tasks at the early stage. Then, the generalized process helps in the construction of a hierarchical generalized knowledge for the specialized groups from bottom-up and encourages the group members to be close together. In the meantime, the specialized learning processes leverage personalized learning to exploit their biased datasets by incorporating higher-level generalized group knowledge from top-level to lower-level groups. In doing so, the group members deviate from the common generalized knowledge. After that, the hierarchical structure will be updated according to the new learning models. 
    
    In the next section, we develop a democratized learning design that results in a hierarchical generalized learning problem. To that end, we propose a novel democratized learning algorithm, \DemLearn to realize as an initial implementation of Dem-AI philosophy.

\section{Democratized Learning Design} \label{S:Learning Design}
Dem-AI philosophy and guidelines in \cite{Dem-AI} envision different designs for a variety of applications and learning tasks. In this work, we focus on developing a novel distributed learning algorithm that consists of the following hierarchical clustering, hierarchical generalization, and learning mechanisms with a common learning task for all learning agents. 
	
\subsection{Hierarchical Clustering Mechanism}
    To construct the hierarchical structure of the Dem-AI system with relevant specialized learning groups, we adopt the commonly used agglomerative hierarchical clustering algorithm (i.e., dendrogram implementation from scikit-learn \cite{hierarchical-clustering-review,scikit-learn}), based on the similarity or dissimilarity of all learning agents. The dendrogram method is used to examine the similarity relationships among individuals and is often used for cluster analysis in many fields of research. During implementation, the dendrogram tree topology is built-up by merging the pairs of agents or clusters having the smallest distance between them, following the bottom-up scheme. Accordingly, the measured distance is considered as the differences in the characteristics of learning agents (e.g., local model parameters or gradients of the learning objective function). Since we obtain a similar performance implementing clustering based on model parameters or gradients, in what follows, we only present a clustering mechanism using the local model parameters. Additional discussion for gradient-based clustering is provided in the supplementary material.
    
    Given the local model parameters $\boldsymbol{\boldsymbol{w}_n}= (\boldsymbol{w}_{n,1},\dots,\boldsymbol{w}_{n,M})$ of learning agent $n$,
    where $M$ is the number of learning parameters, the measure distance between two agents $\boldsymbol{\phi}_{n,l}$ is derived based on the Euclidean distance such as $\boldsymbol{\phi}_{n,l} = \|\boldsymbol{\boldsymbol{w}_n} - \boldsymbol{\boldsymbol{w}_l}\|$.    
     In addition, we consider the average-linkage method \cite{dasgupta2005performance} for distance calculation between an agent and a cluster using the Euclidean distance between the model parameters of the agent and the average model parameters of the cluster members. Accordingly, the hierarchical tree structure is in the form of a binary tree with many levels. In consequence, it will require unnecessarily high storage and computational cost to maintain and be also an inefficient way to maintain a large number of low-level generalized models for small groups. As a result, we keep only the top $K$ levels in the tree structure and discard the lower-levels structure. Therefore, at the top level $K$, the system could have two big groups that have a large number of learning agents.

\subsection{Hierarchical Generalization and Learning Mechanism}
    The $K$ levels hierarchical structure emerges via agglomerative clustering. Accordingly, the system constructs $K$ levels of the generalization, as in Fig~\ref{F:Dem_AI_Principle}. As such, we propose hierarchical generalized learning problems (\textbf{HGLP}) to build these generalized models for specialized groups in a recursive form, starting from the global model $\boldsymbol{w}^{(K)}$ construction at the top level $K$ as follows:
    \textbf{\textbf{HGLP} problem at level $K$} 
	\begin{align}
 	&\underset{\boldsymbol{W}^{(K)}} {\min} \; {L}^{(K)} =  \sum_{i \in \mathcal{S}_K} \frac{N_{g,i}^{(K-1)}}{N_g^{(K)}} {L}_{i}^{(K-1)}(\boldsymbol{w}^{(K-1)}_i|\mathcal{D}^{(K-1)}_{i})\\
    & \qquad\qquad s.t. \quad \boldsymbol{w}^{(K)} = \boldsymbol{w}^{(K-1)}_i, \; \forall i \in \mathcal{S}_K; \label{EQ:GK}
	\end{align}
	where $\boldsymbol{W}^{(K)} = (\boldsymbol{w}^{(K)},  \boldsymbol{w}^{(K-1)}_1,\dots,\boldsymbol{w}^{(K-1)}_{|\mathcal{S}_K|})$, 
	$\mathcal{S}_K$ is the set of subgroups of the top level group, ${L}_i^{(K-1)}$ is the loss function of subgroup $i$ given its collective dataset $D_{i}$. The objective function is weighted by a fraction of the number of learning agents $N_g^{(K-1)}$ of the subgroup $i$, and the total number of learning agents $N_g^{(K)}$ in the system. Hence, the subgroups which have more learning agents have higher impact to the generalized model at level $K$. The hard constraints in (\ref{EQ:GK}) enforce these subgroups to share a common learning model (i.e., a global variable $\boldsymbol{w}^{(K)}$). To preserve the specialization capabilities of each subgroup, these constraints (\ref{EQ:GK}) could be relaxed by using additional proximal terms in the objective. In this way, the problem encourages the subgroup learning models to become close to the global model but not necessarily equal. Thus, the relaxed problem \textbf{HGLP'} is defined as follows:
	
    \textbf{\textbf{HGLP'} problem at level $K$}
 	\begin{align}
 	\underset{\boldsymbol{W}^{(K)}}{\min} & \sum_{i \in \mathcal{S}_K} \frac{N_{g,i}^{(K-1)}}{N_g^{(K)}}\Big( {L}_{i}^{(K-1)}(\boldsymbol{w}^{(K-1)}_i|\mathcal{D}^{(K-1)}_{i}) \nonumber\\
 	&+ \frac{\mu_K}{2} \|\boldsymbol{w}^{(K)} - \boldsymbol{w}^{(K-1)}_{i}\|^2\Big),
 	\label{EQ:GK1}
	\end{align}
	where $\mu_K$ denotes the trade-off between the learning loss and the generalization constraint enforcing the group learning models to be close to the global model $w^{(K)}$.
	Since the dataset is distributed and only available at the learning agents, the problem \eqref{EQ:GK1} at the top level $K$ can be solved starting from its members problem first. Accordingly, the hierarchical generalized structure is emerged naturally following the bottom-up scheme where the learning models at lower levels are updated before solving the higher level generalized problems of its upper-group. Specifically, the problem \eqref{EQ:GK1} can be decentralized and solved by the following problem of each subgroup $i$ at the level $K-1$.
	


	
   \textbf{\textbf{HGLP} problem for each group $i$ at level $K-1$}
	\begin{align}
	    \underset{\boldsymbol{W}^{(K-1)}} {\min}
	    &  \sum_{j \in \mathcal{S}_{i, K-1}} \frac{N_{g,j}^{(K-2)}}{N_{g}^{(K)}} \Big( {L}_{j}^{(K-2)}(\boldsymbol{w}^{(K-2)}_j|\mathcal{D}^{(K-2)}_{j}) \nonumber \\
	    &+  \frac{\mu_{K-1}}{2} \|\boldsymbol{w}^{(K-2)}_{j} - \boldsymbol{w}_i^{(K-1)}\|^2 \Big) \nonumber \\
	    &+ \frac{\mu_{K} N_{g,i}^{(K-1)}}{2 N_g^{(K)}} \|{\boldsymbol{w}^{(K)} - \boldsymbol{w}^{(K-1)}_{i}}\|^2; \notag
	\end{align}

	where $\boldsymbol{W}^{(K-1)} = (\boldsymbol{w}_i^{(K-1)}, \boldsymbol{w}^{(K-2)}_1,\dots,\boldsymbol{w}^{(K-2)}_{|\mathcal{S}_{i,K-1}|}).$
	Therefore, we make a general approximation form of the generalized learning problem for the group $i$ at the level $k$ given the prior higher generalized model $\boldsymbol{w}^{(k+1)}$ as follows:
	
	\textbf{HGLP problem for each group $i$ at level $k$}
	\begin{align}
	    \label{EQ:level_k_prob}
	    \underset{\boldsymbol{W}^{(k)}} {\min}
	    & \sum_{j \in \mathcal{S}_{i, k}} \frac{N_{g,j}^{(k-1)}}{N_{g}^{(K)}}\Big( {L}_{j}^{(k-1)}(\boldsymbol{w}^{(k-1)}_j|\mathcal{D}^{(k-1)}_{j}) \nonumber\\&
	    + \frac{\mu_{k}}{2} \|\boldsymbol{w}^{(k-1)}_j-\boldsymbol{w}_i^{(k)}\|^2 \Big)
	   + \frac{\mu_{k+1} N_{g,i}^{(k)}}{2 N_g^{(K)}} \|\boldsymbol{w}^{(k+1)}-\boldsymbol{w}^{(k)}_{i} \|^2,
	\end{align}
\begin{figure}[t]
	\centering
	\includegraphics[width=0.5\linewidth]{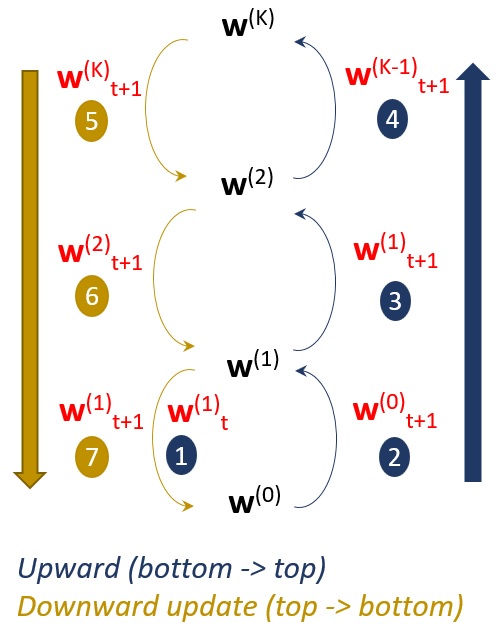}
	\caption{Hierarchical update the generalized models.}
	\label{F:Recursive}
\end{figure}
where $\boldsymbol{W}^{(k)} = (\boldsymbol{w}_i^{(k)}, \boldsymbol{w}^{(k-1)}_1,\dots,\boldsymbol{w}^{(k-1)}_{|\mathcal{S}_{i,k}|})$,   $N_{g,i}^{(k)}$ is the number of learning agents of group $i$ and $\boldsymbol{w}^{(k+1)}$ the learning model of the upper-group at level $k+1$ in which group $i$ belongs. Since there exists coupling between the upper and lower levels, and the training dataset is decentralized, the learning problem (\ref{EQ:level_k_prob}) of the group $i$ at level $k$ cannot be solved directly. Therein, similar to FL, the learning loss of the group can be distributed amongst the group members \cite{mcmahan2016communication}. As a result, the objective of the group problem has the remaining proximal terms forcing the learning models in different levels to be close to each other. Therefore, the learning model is constructed with the model of upper-group at level $k+1$ and group members at level $k-1$ models by solving the following problem:

	\begin{align}
	    \label{EQ:level_k_prob2}
	    \underset{\boldsymbol{w}} {\min}
	    & \sum_{j \in \mathcal{S}_{i, k}} \frac{\mu_{k}N_{g,j}^{(k-1)}}{N_{g}^{(K)}}
	     \|\boldsymbol{w}^{(k-1)}_j-\boldsymbol{w}\|^2  \nonumber\\
	   &+ \frac{\mu_{k+1} N_{g,i}^{(k)}}{ N_g^{(K)}} \|\boldsymbol{w}^{(k+1)}- \boldsymbol{w}\|^2. 
	\end{align}
	
	 The closed form of the optimal solution of the problem (\ref{EQ:level_k_prob2}) can be handily derived by setting the gradient to zero as follows:
	 
    \begin{align}
        \sum_{j \in \mathcal{S}_{i, k}} \mu_{k}N_{g,j}^{(k-1)}(\boldsymbol{w}^{(k-1)}_j-\boldsymbol{w}^*) = 
        \mu_{k+1}N_{g,i}^{(k)} (\boldsymbol{w}^* -\boldsymbol{w}^{(k+1)} ) \nonumber
    \end{align}
    
    Thus, the learning model of group $i$ can be updated as

    
    \begin{equation}
        \boldsymbol{w}_i^{(k)} = \alpha \boldsymbol{w}^{(k+1)} + (1-\alpha)\sum_{j \in \mathcal{S}_{i, k}} \frac{N_{g,j}^{(k-1)}}{N_{g,i}^{(k)}} \boldsymbol{w}^{(k-1)}_j, \label{EQ:closed_form}
    \end{equation}
    where $\alpha= \mu_{k+1} / (\mu_k+\mu_{k+1})$. The trade-off parameter $\alpha$ can be tuned later in the experiments to control the contribution from the learning models of upper-group and group members.

    Given the closed form solution (\ref{EQ:closed_form}), the coupling between different levels can be approximated by splitting the model updates at each level via the bottom-top, and then, top-bottom scheme. In Fig.\ref{F:Recursive}, we show the illustration of the proposed hierarchical updates. In particular, the lower groups and members are updated first before the upper-groups. And then, the updated upper-groups broadcast the parameters to its group members to finish one update cycle.

	At the lowest level, each learning agent $n$ can actually perform the local training process to fit its private data with the personalized learning problem using the latest hierarchical generalized models as follows:

    \textbf{\textbf{PLP} at level $0$} 

	\begin{align}
	\label{EQ:LPL_level_0}
	\boldsymbol{w}^{(0)}_n 
	&=\underset{\boldsymbol{w} \in \mathcal{W}}{\arg \min}~L^{(0)}_n (\boldsymbol{w}|\mathcal{D}^{(0)}_n) + \frac{\mu}{2}\|\boldsymbol{w}-\boldsymbol{w}_n^{(1)}\|^2,  
	\end{align}
	
    where $L^{(0)}_n$ is the personalized learning loss function for the learning task (e.g., cross entropy loss for classification \cite{cross_entropy}) given its personalized dataset $\mathcal{D}^{(0)}_n$, $N_{n,g}^{(k)}$ is the number of learning agents of the level-$k$ group in which the agent $n$ belongs. Solving the \textbf{PLP} problem, the learning agent can update their personalized model ${w}^{(0)}_n$ belonging to the parameterized deep learning model set $\mathcal{W}$. In this personalized level, the number of group member is $1$.
	
	\begin{algorithm}[t]
		\caption{ Democratized Learning (\DemLearn)} 
		\label{DemLearn} 
		\begin{algorithmic}[1]
            \State \textbf{Input:} $K, T, \tau.$
            \For{ $t=0,\dots,T-1$}
        		\For{learning agent $n=1,\dots,N  $}
        		   \State \begin{varwidth}[t]{0.9\linewidth}Agent $n$ uses the upper-group model $\boldsymbol{w}^{(1)}_{n,t}$ as the initial model;
        		   \end{varwidth}
    		      \State \begin{varwidth}[t]{0.9\linewidth} 
        		    Agent $n$ iteratively updates the personalized learning model $\boldsymbol{w}^{(0)}_{n,t+1}$ as an in-exact minimizer  
        		    (i.e., \textit{gradient based}) of the following problem:
        		    \begin{align}
        		         &\underset{\boldsymbol{w} \in \mathcal{W}}{\min}~ L^{(0)}_n (\boldsymbol{w}|\mathcal{D}_n) + 
        		        \frac{\mu}{2} \|\boldsymbol{w}-\boldsymbol{w}^{(1)}_{n,t}\|^2; 
        		        \label{EQ:mu_PLP}
        		    \end{align}
        		    \end{varwidth}
    		        \State Agent $n$ sends updated learning model to the server; 
    		   \EndFor
    		   \If {$(t \mod \tau = 0)$}
        		   \State \begin{varwidth}[t]{0.9\linewidth}Server reconstructs the hierarchical structure by the clustering algorithm;\end{varwidth}
        		        
      		  \EndIf
  		      \State {\begin{varwidth}[t]{0.9\linewidth} \textbf{Hierarchical Update:}
  		      Each group $i$ at each level $k$ performs an update for its learning model from bottom to top for updating the contribution of group members:
  		      \begin{equation}
  		       \boldsymbol{w}_{i,t+1}^{(k)} =  \sum_{j \in \mathcal{S}_{i, k}} \frac{N_{g,j}^{(k-1)}}{N_{g,i}^{(k)}} \boldsymbol{w}^{(k-1)}_{j,t+1}.
  		       \label{EQ:bottom-top}
  		       \end{equation}

  		      After the top-level model is updated, the lower-levels starts updating top to bottom for the contribution of the upper-group as follows:
  		            \begin{equation}
                        \boldsymbol{w}_{i,t+1}^{(k)} = \alpha \boldsymbol{w}_{t+1}^{(k+1)} + (1-\alpha) \boldsymbol{w}_{i,t+1}^{(k)}. 
                        \label{EQ:top-bottom}
                        \end{equation}
                
            The updated learning models at level $1$ (i.e., $\boldsymbol{w}^{(1)}_{t+1}$) are then broadcast to all agents to update their local models following equation (\ref{EQ:top-bottom}).
            \end{varwidth}
                    
      		  }
  		   \EndFor
		\end{algorithmic}
	\end{algorithm}	
	
\begin{figure*}[t]
    \centering
    \begin{subfigure}{\linewidth}
        \centering
        \includegraphics[width=\linewidth]{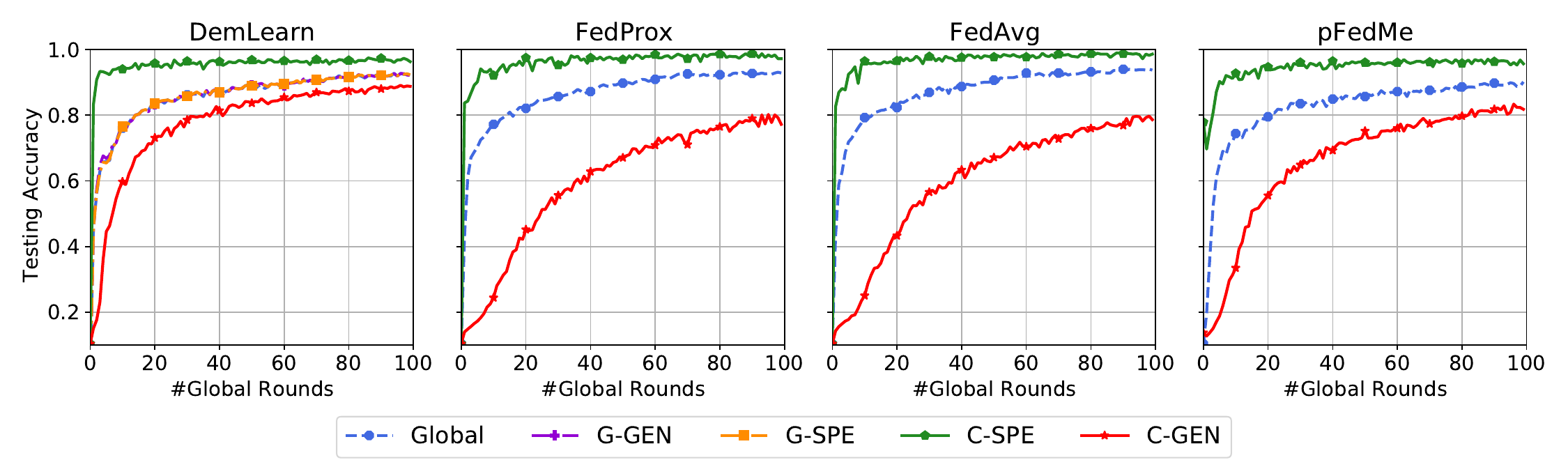}
    	\caption{Experiment with MNIST dataset.}
    	\label{Fig:avg_prox_fed1a}
	\end{subfigure}
    \begin{subfigure}{\linewidth}
        \centering
        \includegraphics[width=\linewidth]{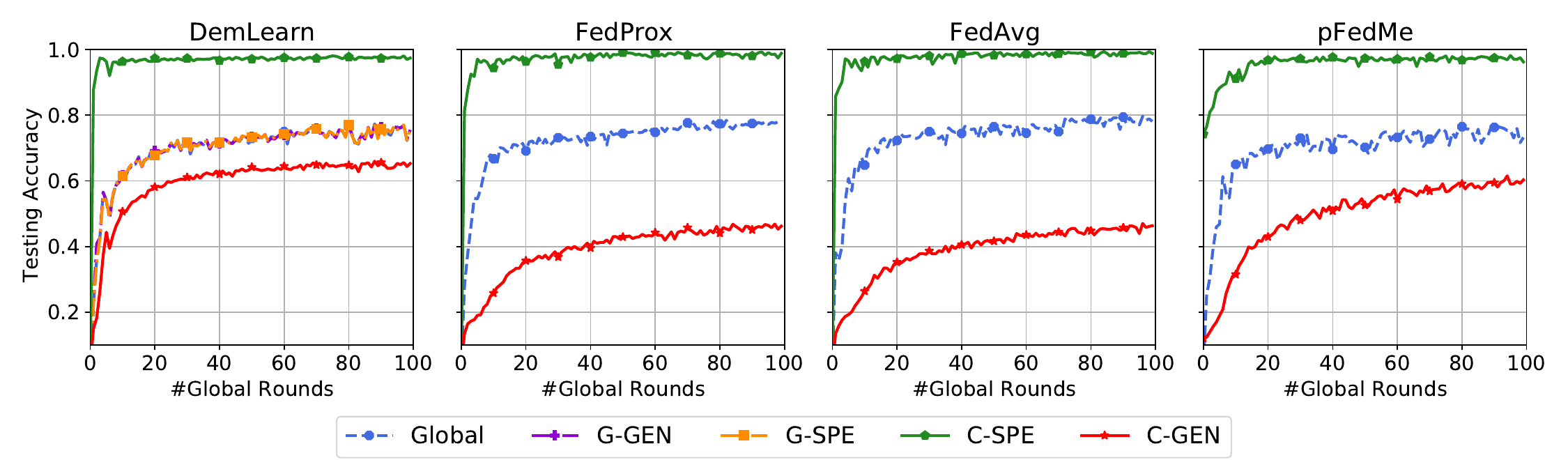}
    	\caption{Experiment with Fashion-MNIST dataset.}
    	\label{Fig:avg_prox_fed1b}
	\end{subfigure}
    \begin{subfigure}{\linewidth}
        \centering
        \includegraphics[width=\linewidth]{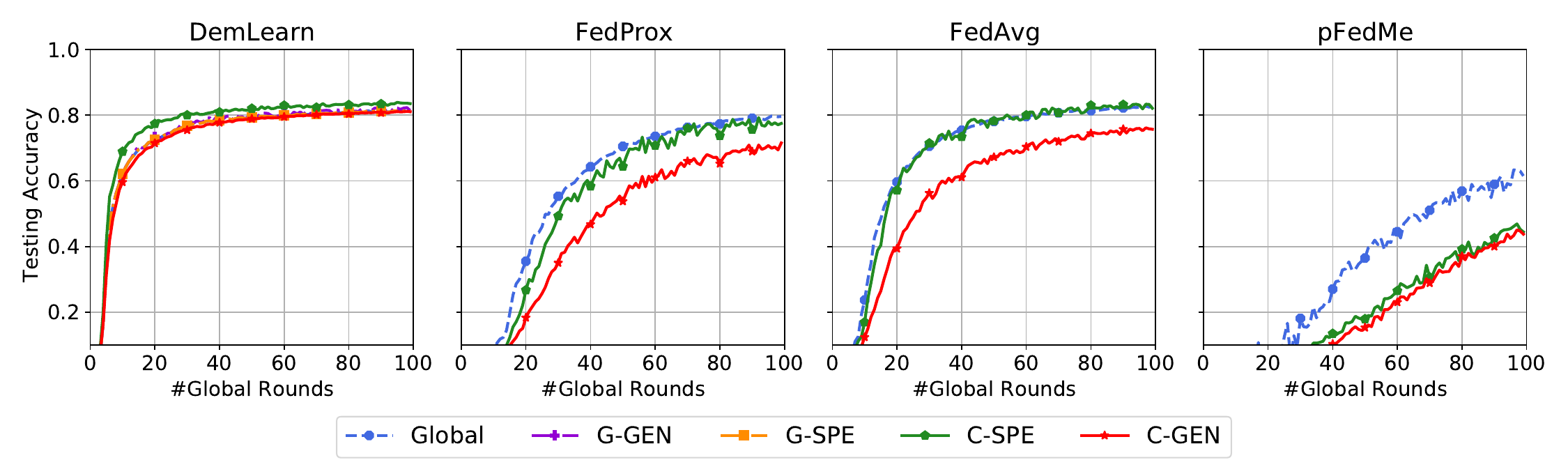}
    	\caption{Experiment with Federated Extended MNIST dataset.}
    	\label{Fig:avg_prox_fed1c}
	\end{subfigure}
    \caption{Performance comparison of \DemLearn versus FedAvg, FedProx, and pFedMe.}
	\label{Fig:avg_prox_fed1}
\end{figure*}
\begin{figure*}[t]
        \centering
        \includegraphics[width=0.76\linewidth]{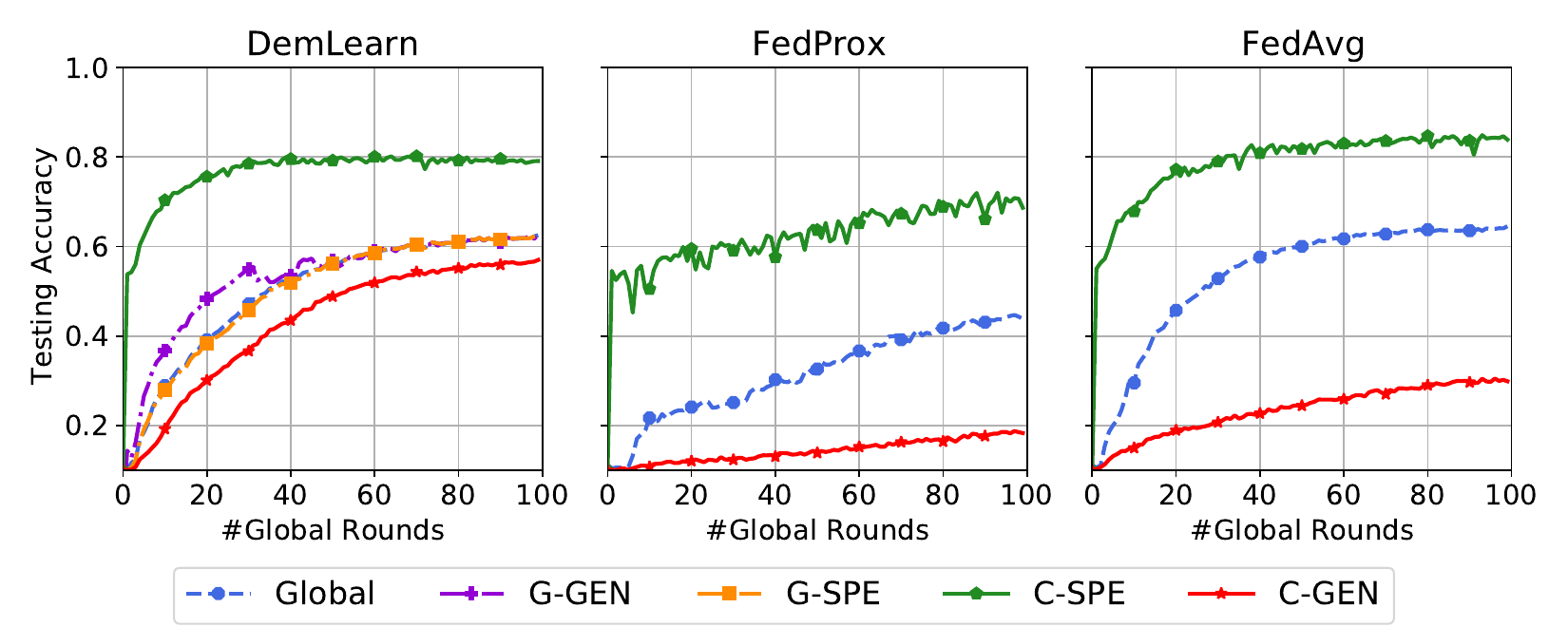}
	\caption{Performance comparison of \DemLearn versus FedAvg, FedProx with CIFAR-10 dataset.}
	\label{Fig:avg_prox_fed2}
\end{figure*}
\setlength{\textfloatsep}{1pt}

	\subsection{Democratized Learning Algorithm}
     Inspired by the FedAvg \cite{mcmahan2016communication} and FedProx \cite{FedProx2020} algorithms, we adopt the aforementioned recursive analysis and hierarchical clustering mechanism to develop a novel democratized learning algorithm, namely \DemLearn. The details of our proposed algorithm are presented in Alg. \ref{DemLearn}. 
     Each agent $n$ uses the upper-group model at level $1$ (i.e., $\boldsymbol{w}^{(1)}_{n,t}$) as the initial learning model. Thereafer, the agent iteratively solves the \textbf{PLP} problem in the equation \eqref{EQ:mu_PLP} based on the gradient method. The updated client model will be sent to the central server to perform hierarchical clustering and update from the generalized level $1$ to the level $K$.
    After every $\tau$ global rounds, the hierarchical structure is reconstructed according to the changes in the personalized learning model of agents. 
    The generalized learning models of groups are updated, respectively, in bottom-top, and then top-bottom fashion, following the equations (\ref{EQ:bottom-top}) and (\ref{EQ:top-bottom}) as the approximation of the closed form solution (\ref{EQ:closed_form}). This allows the lower level subgroups to contribute their knowledge for updating the group model. In return, they receive (and incorporate) the better generalized knowledge from the upper-groups that enhances the generalization capacity of their local learning models.
    Additionally, we introduce an amplification trick in the bottom-top update for the first $5$ rounds to speed up the initial stage of learning process. Accordingly, the update from group members (i.e., $\boldsymbol{w}^{(k)}_{t+1}$ in the equation (\ref{EQ:bottom-top})) is multiplied with a scaling constant $1.15$ in the first $5$ rounds.

\begin{figure*}[t]
    \centering
    \begin{subfigure}{\linewidth}
        \centering
    	\includegraphics[width=\linewidth]{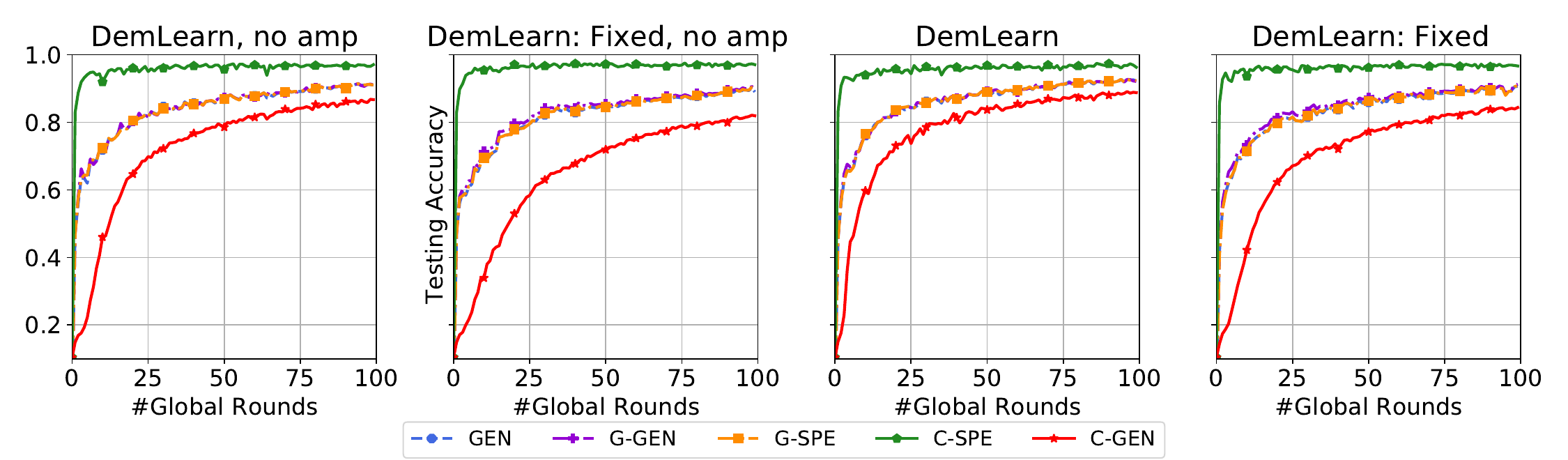}
    	\caption{Experiment with MNIST dataset.}
	\label{Fig:avg_K_levels1}
	\end{subfigure}
    \begin{subfigure}{\linewidth}
        \centering
    	\includegraphics[width=\linewidth]{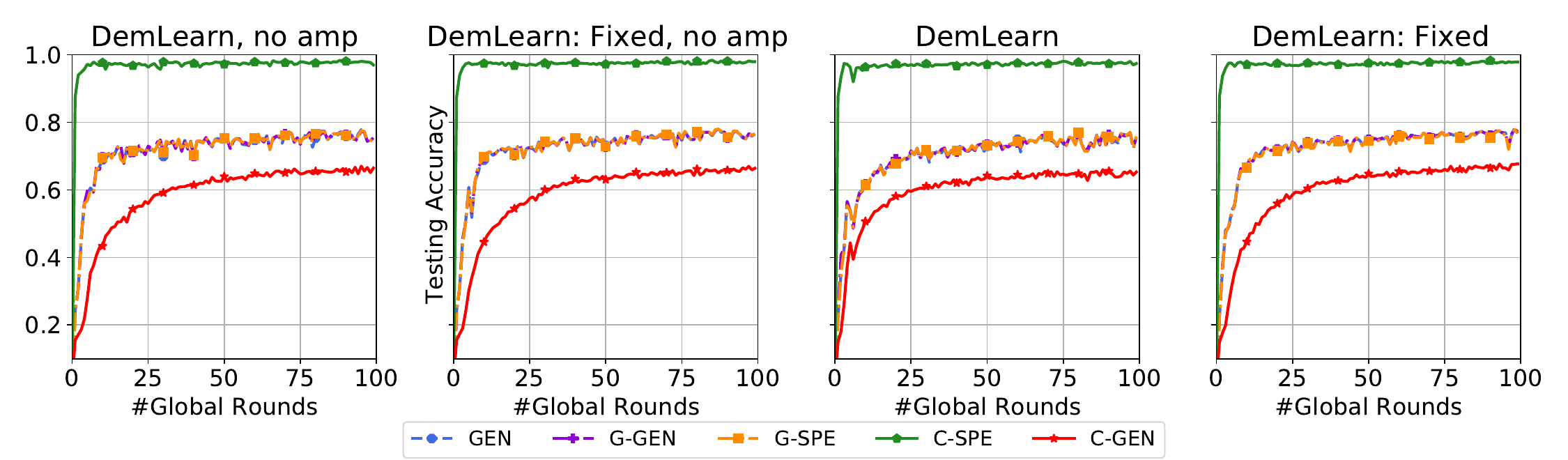}
    	\caption{Experiment with Fashion-MNIST dataset.}
    	\label{Fig:avg_K_levels2}
	\end{subfigure}
	\caption{Comparison of algorithms for a fixed and a self-organizing hierarchical structure.}
	\label{Fig:avg_K_levels}
\end{figure*}


\begin{figure*}[t]
    \centering
    \begin{subfigure}{\linewidth}
        \centering
    	\includegraphics[width=\linewidth]{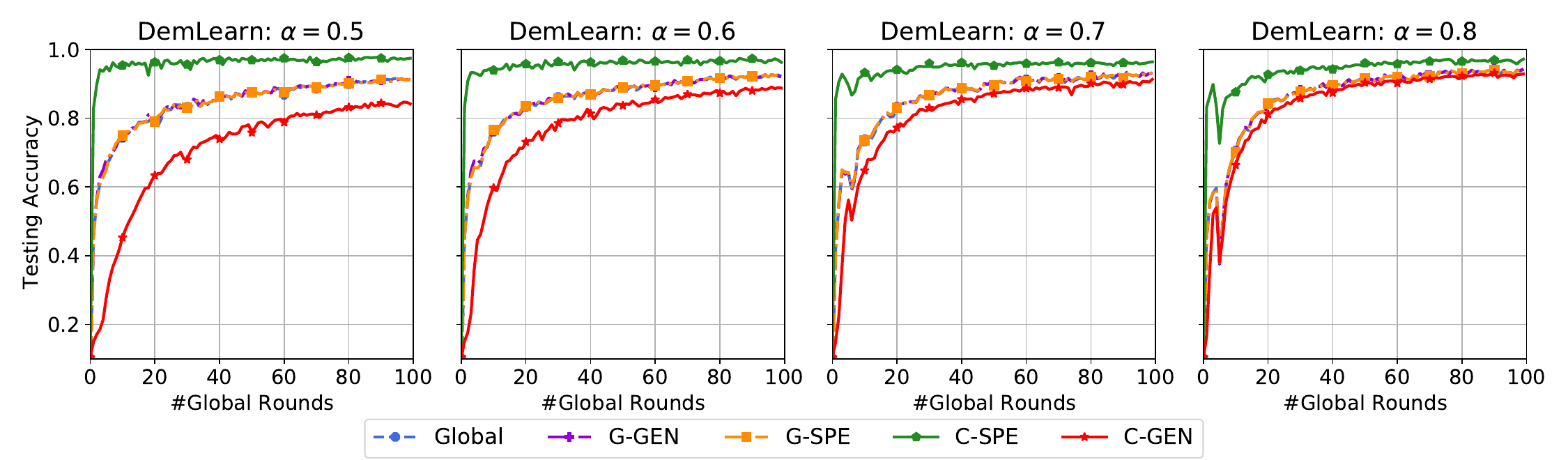}
    	\caption{Experiment with MNIST dataset.}
    	\label{Fig:alpha_figa}
	\end{subfigure}
    \begin{subfigure}{\linewidth}
        \centering
    	\includegraphics[width=\linewidth]{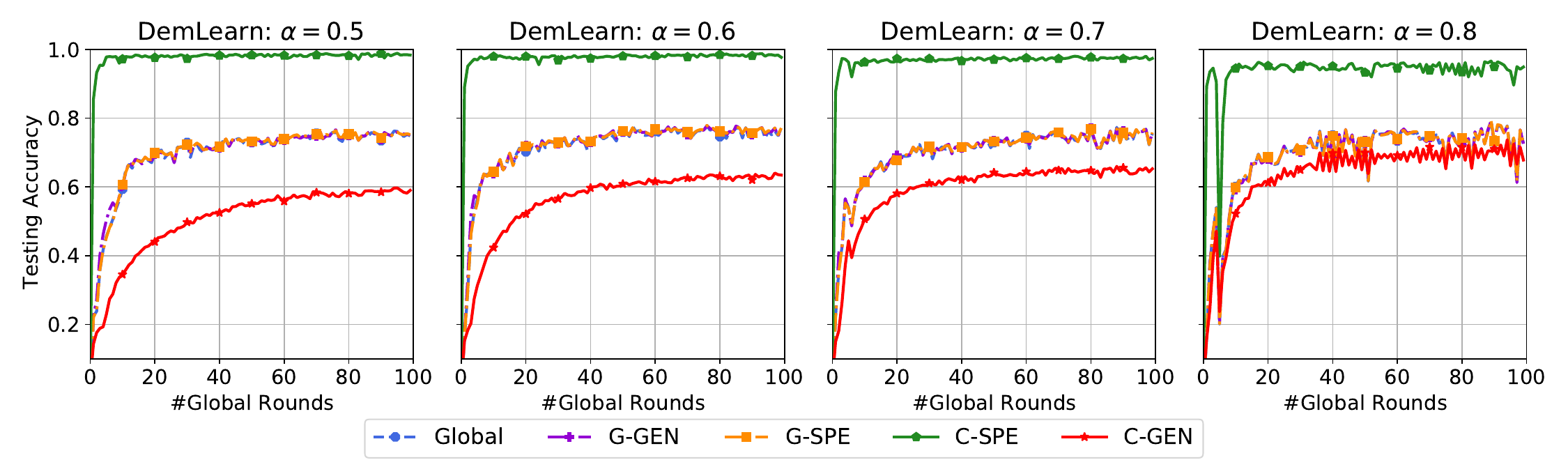}
    	\caption{Experiment with Fashion-MNIST dataset.}
    	\label{Fig:alpha_figb}
	\end{subfigure}
	\caption{Performance of \DemLearn by varying $\alpha$ with MNIST and Fashion-MNIST datasets.}
	\label{Fig:alpha_fig}
\end{figure*}

\section{Experiments}\label{S:Performance Evaluation}
\subsection{Setting} 
In this section, we validate the efficacy of the \DemLearn algorithm with the MNIST \cite{lecun1998mnist}, Fashion-MNIST \cite{xiao2017fashion_mnist}, Federated Extended MNIST \cite{caldas2018leaf}, and CIFAR-10 \cite{krizhevsky2009learning} datasets for handwritten digits and fashion images recognition, and objects recognition, respectively.
 We conduct the experiments with $50$ clients, where each client has median numbers of data samples at $64$, $70$ and $785$ with MNIST, Fashion-MNIST dataset, and CIFAR-10, respectively. Different from these three datasets, we also experiment with the FE-MNIST dataset which has more number of classes such as $10$ digits and $25$ lowercase and $25$ uppercase. Accordingly, we select $50$ clients from $3559$ total of users which have at least $50$ data samples in FE-MNIST dataset. Using these datasets, $20\%$ of data samples on each client are used for evaluating the model testing performance. We divide the total dataset such that each client has a small amount of data from two specific labels amongst the overall ten in both datasets. In doing so, we replicate a scenario of biased personal datasets, i.e., highly unbalanced data and a small number of training samples can be collected at agents. The learning models consist of two convolution layers followed by two pooling and two fully connected layers whereas three convolution layers are used in the CIFAR-10 dataset. We set the update period $\tau=1$, and validate the performance of the proposed algorithm with $K=4$ generalized levels. Our implementation is developed based on the available code of FedProx in \cite{FedProx2020}. \DemLearn, FedAvg and FedProx use the common learning rate $\eta=0.05$, local epoch $E=2$, batch size $B=10$. For FedProx, we set the parameter $\mu=0.5$. Meanwhile, pFedMe needs detailed tuning to obtain a competitive accuracy for different datasets.
The Python implementation of our proposed algorithm using Pytorch and datasets are available at \url{https://github.com/nhatminh/Dem-AI}. \par

\subsection{Results}
Existing FL approaches such as FedProx and FedAvg focus more on the learning performance of the global model rather than the learning performance at clients. Therefore, for forthcoming personalized applications, we implement \DemLearn and measure the learning performance of all clients and the group models. In particular, we conduct evaluations for specialization (C-SPE) and generalization (C-GEN) of learning models at agents on average that are defined as the performance in their local test data only, and the collective test data from all agents in the region, respectively. Accordingly, we denote Global as global model performance; and G-GEN, G-SPE are the average generalization and specialization performance of group models, respectively. In addition to the standard C-SPE performance for local models, the introduced C-GEN performance is an important metric that shows the generalized capabilities of local models. Even though the biased local models can achieve high C-SPE values from very early, particularly due to their small local datasets, they still have a very low generalized capabilities which can help to produce good predictions according to the frequent changes of users. Meanwhile, the global and group models have the highest generalized capabilities, but a lower specialized capabilities during deployment at the clients.




\indent In Fig.~\ref{Fig:avg_prox_fed1},~\ref{Fig:avg_prox_fed2}, we conducted performance comparisons of our proposed methods, \DemLearn with the three FL methods, FedAvg \cite{mcmahan2016communication}, FedProx \cite{FedProx2020}, and pFedMe \cite{canh2020personalized} as baselines on four benchmark datasets, MNIST, Fashion-MNIST, FE-MNIST and CIFAR-10. 
Fig.~\ref{Fig:avg_prox_fed1a} depicts the performance comparison of \DemLearn with FedProx and FedAvg with the MNIST dataset. Experimental evaluations show that the proposed approach outperforms the baselines in terms of the convergence speed, especially to obtain better client generalization performance. We observe that the local model requires only $40$ rounds to reach the C-GEN performance level of $80\%$ using the proposed algorithm, whereas existing FL algorithms such as FedProx, FedAvg, and pFedMe take more than $80$ global rounds to achieve a near-competitive level of performance as ours. Furthermore, after $100$ rounds, \DemLearn obtains better average client generalization performance (i.e., $88.77\%$) accross client models and comparable C-SPE and Global performance as of FedAvg and pFedMe.


Following similar trends, Fig.~\ref{Fig:avg_prox_fed1b}, Fig.~\ref{Fig:avg_prox_fed1c}, Fig.~\ref{Fig:avg_prox_fed2} depict that \DemLearn algorithm perform better than FedProx and FedAvg with the Fashion-MNIST, FE-MNIST and CIFAR-10 datasets in terms of the client generalization performance. In Fig.~\ref{Fig:avg_prox_fed1c}, the pFedMe algorithm found difficulties in tuning parameters to obtain a comparable performance and showed lower performance and more fluctuated rather than that of the other algorithms for the FE-MNIST dataset. Thus, pFedMe obtains a slow improvement of the client models in both specialization and generalization. Meanwhile, our proposed algorithm exhibits stable convergence speed and efficiency to achieve consistently high performance of learning models at all levels. In Fig.~\ref{Fig:avg_prox_fed2}, we observe \DemLearn suffers a  slight degradation in C-SPE and Global performance to gain a high C-GEN performance. After $100$ rounds, \DemLearn demonstrates good trade-off learning capabilities of client models with high C-SPE ($79.09\%$) and C-GEN ($57\%$) performance, while other baseline algorithms only produce biased local models with low generalized capabilities.

 In Fig.~\ref{Fig:avg_K_levels}, we evaluate and compare the performance of the proposed algorithm for a fixed and a self-organizing hierarchical structure via periodic reconstruction (i.e., $\tau=1$). We observe that \DemLearn benefits from the self-organizing mechanism, and can provide slightly better generalization capability of client models. In addition, the amplification in the first $5$ rounds help the proposed algorithm can speeds up the initial performance of the \DemLearn algorithm.

\subsection{Tuning Parameters of the proposed algorithm} 
In this subsection, we show the impact of parameter $\alpha$ in the testing accuracy of MNIST and Fashion-MNIST datasets, as shown in Fig. \ref{Fig:alpha_fig}. As can be seen in the results, when $\alpha$ is large, we obtain high performance of the client generalization while slight degradation in their specialized capabilities. As such, increasing the value of $\alpha$ enhances generalization, but also reduces the specialization performance of client models to a marginal extent. Since $\alpha=\mu_{k+1} / (\mu_k+\mu_{k+1})$ controls the contribution from the learning models of upper-group and group members, tuning $\alpha$ produces the impact in disseminating the generalized knowledge from upper-groups to the lower level groups and clients. At each level $k$, the higher value of $\alpha$ signifies the objective is more focused on minimizing the gap with the upper-group model at $k+1$ than the lower-group models at $k-1$. Furthermore, we evaluate the impact of other parameters, such as $\mu,K,\tau$, but they do not show any clear effects on the performance of \DemLearn, likely due to the small-scale simulation settings. In addition to which, reducing the frequency of cluster updates by increasing the parameter $\tau$ can help the algorithm obtain comparable performance with a lower reorganizing cost of the hierarchical structure. However, due to the limited scope of the simulated datasets and settings, we advocate the impact of defined parameters on the performance of the proposed algorithm is better realized when evaluating the algorithm with a more practical/experimental data. In particular, we keep it as our future work to evaluate our algorithm on dataset that has a hierarchical structure based on groups of similar users, or exceedingly large and different users.


\begin{figure*}[t]
    \centering
    \begin{subfigure}{\linewidth}
    	\includegraphics[width=\linewidth]{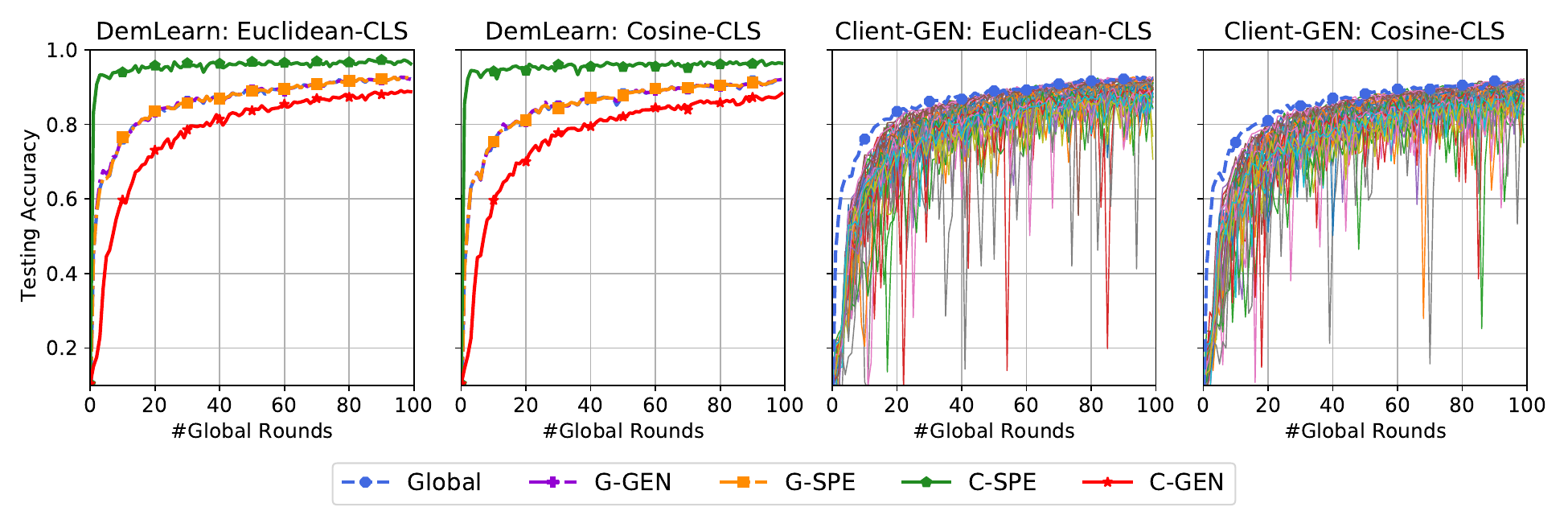}
    	\caption{Experiment with MNIST dataset.}
    \end{subfigure}
    \begin{subfigure}{\linewidth}
    	\includegraphics[width=\linewidth]{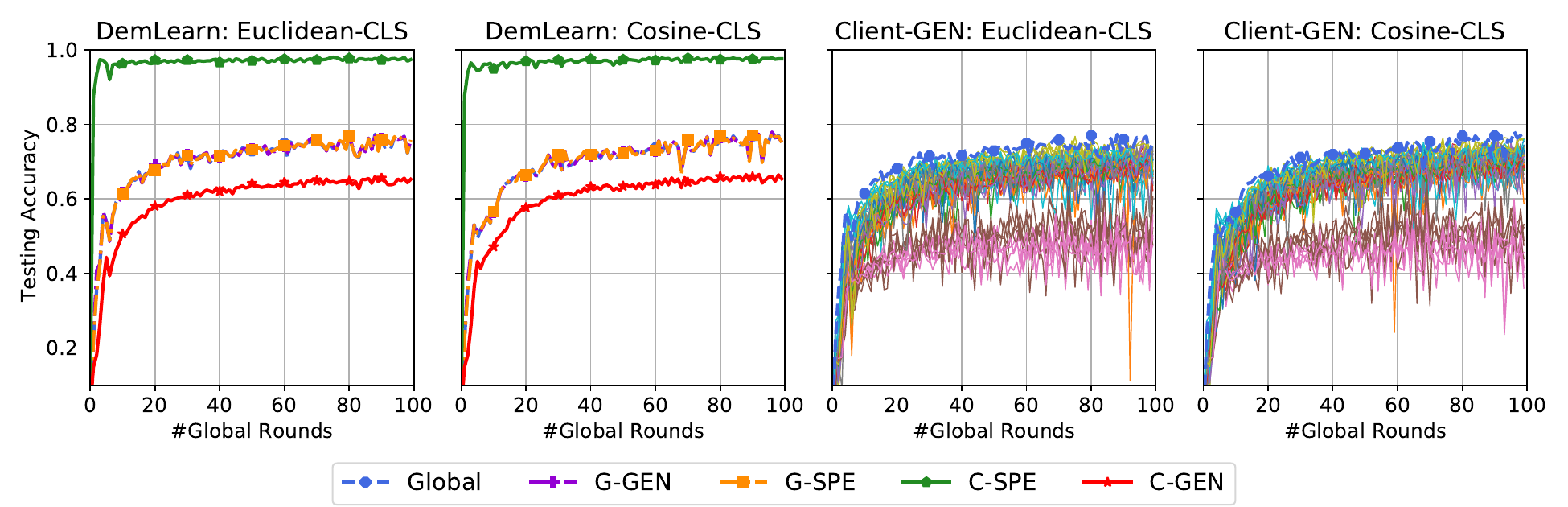}
    	\caption{Experiment with Fashion-MNIST dataset.}
    \end{subfigure}
    \caption{Comparison of different metrics using in hierarchical clustering.}
    \label{Fig:w_vs_g}
\end{figure*}

\begin{figure*}[t]
	\begin{subfigure}{\linewidth}
    	\includegraphics[width=\linewidth]{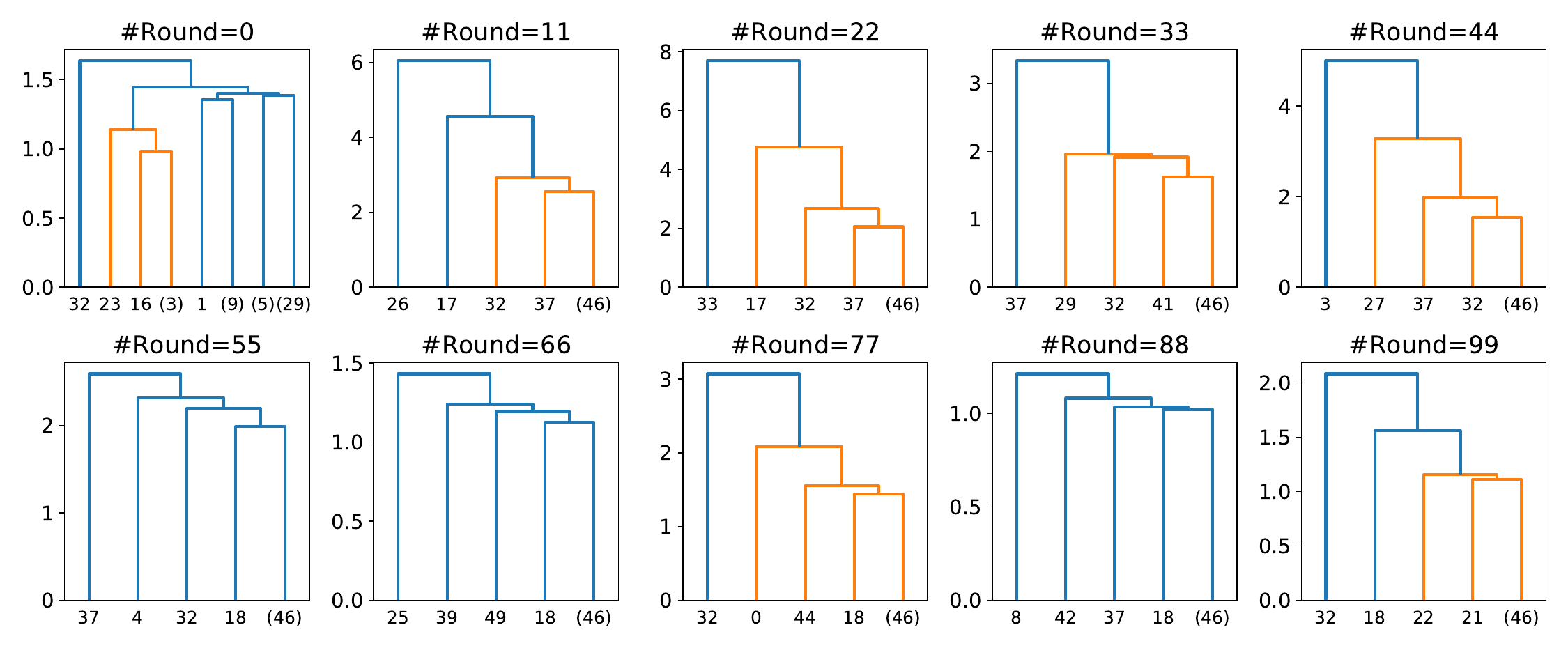}
    	\caption{Hierarchical clustering based on Euclidean distance}
	\end{subfigure}
	\begin{subfigure}{\linewidth}
    	\includegraphics[width=\linewidth]{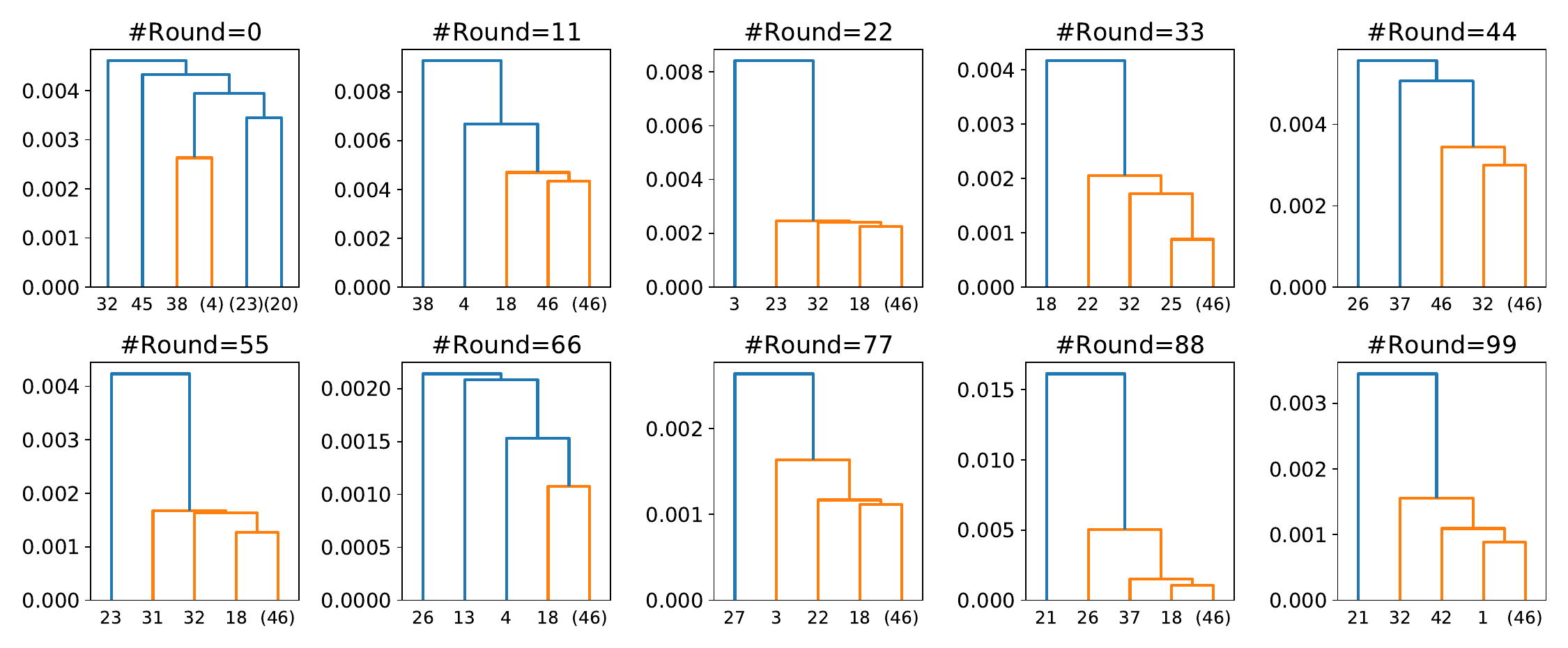}
    	\caption{Hierarchical clustering based on Cosine similarity distance}
    \end{subfigure}
    \caption{Topology changes via hierarchical clustering in \DemLearn algorithm with MNIST dataset.}
    \label{Fig:cluster_topology}
\end{figure*}

\subsection{Clustering Approach}
In addition to the euclidean distance between learning parameters in the hierarchical clustering algorithm, we can also evaluate the measured distance $\phi^{(cos)}_{n,l}$ between two learning models based on the cosine similarity as follows:

	\begin{equation}
    	\phi^{(cos)}_{n,l} = \cos(\boldsymbol{w}_n,\boldsymbol{w}_l) = \frac{\sum_{m=1}^{M}\boldsymbol{w}_{n,m}\boldsymbol{w}_{l,m}}{\sqrt{\sum_{m=1}^{M}\boldsymbol{w}_{n,m}^2}\sqrt{\sum_{m=1}^{M}\boldsymbol{w}_{l,m}^2}}.
	\end{equation}


Our experimental results demonstrate the clients and groups models show almost similar learning performance (Fig.~\ref{Fig:w_vs_g}) and different trends of cluster topology (Fig.~\ref{Fig:cluster_topology}) using \DemLearn algorithm. As shown in the Client-GEN subfigure, some outliers contribute to the drop in client-GEN performance throughout the training. At the same time, most clients have comparable generalized capacities with the global model. Accordingly, the illustration further helps us figure out the exceedingly different clients with low C-GEN performance. That means, in each round, the outliers are detected (and classified) by a clustering mechanism that is exceedingly different from other client model parameters. However, we note that the outliers are not always the same and change throughout their local model updates. But some appear several times that provide meaningful information for the learning system. Furthermore, as shown in Fig.\ref{Fig:cluster_topology}, the difference in cluster topology does not affect much on the overall learning performance.

\subsection{Discussion}
Compared to the FedAvg and FedProx, we have a similar on-device computational cost for solving \textbf{PLP} problem using the gradient descent method. Different from the others, pFedMe requires extra steps for the $\theta$ approximation; hence, requires more computation time on the client device. On the other hand, for the computation time requirement at the server, the \DemLearn algorithm requires extra computation cost for hierarchical clustering and a negligible cost for hierarchical updates. We note that the cost of the hierarchical clustering depends on the model size and the number of learning agents. In principle, the standard algorithm for hierarchical agglomerative clustering has a time complexity of $O(n^3)$ and requires $O(n^2)$ memory \cite{nielsen2016introduction}. In this work, we perform evaluation of the algorithm with $n=50$ users and obtain a running time of $0.0015s$ per step on average when using a computer with CPU i7-7700K and memory of 32 GB. For the communication cost, all algorithms require similar costs to send the model parameters.

In practice, to deploy a typical hierarchical structure the Dem-AI systems can include three entities: 
\begin{itemize}
\item A cloud server that handles the global model (root node) and its sub-level generalized groups;
\item Distributed regional servers, which are edge servers deployed within each region and whose role is to manage the subgroups and learning agents.
\item Learning agents.
\end{itemize}

Our implementation for hierarchical clustering is in centralized manner in this paper. However, the agglomerative hierarchical clustering mechanism merges the learning agents and groups in a bottom-up manner, then it is possible to implement it in a decentralized manner. Therefore, we can operate the hierarchical clustering in the cloud server for top levels and decentralized implementation for regions at edge servers for lower-level groups. Also, this grouping mechanism can mitigate the negative effects in the aggregation of exceedingly different learning agents to construct hierarchical generalized models. In that way, we maintain three or more levels of generalized models instead of a single global model. To the best of our knowledge, this design has not been considered yet in recent studies of personalized FL. 

\section{Conclusion and Future Works}\label{S:Conclusion}
    The novel Dem-AI philosophy has provided general guidelines for specialized, generalized, and self-organizing hierarchical structuring mechanisms in large-scale distributed machine learning systems. Inspired by these guidelines, we have formulated the hierarchical generalized learning problems and developed a novel distributed learning algorithm, \DemLearn.
    In this work, based on the similarity in the learning characteristics, the agglomerative clustering enables the self-organization of learning agents in a hierarchical structure, which gets updated periodically. 
    Detailed analysis of experimental evaluations has shown the advantages and disadvantages of the proposed algorithm. Compared to conventional FL, we show that \DemLearn significantly improves the generalization performance of client models without largely compromising the specialization performance of clients models. As a result, \DemLearn enables good trade-off learning capabilities of client models with high C-SPE and C-GEN performance while other algorithms can only produce biased local models with low generalized capabilities. These observations benefit for a better understanding and improvement in specialization and generalization performance of the learning models in future Dem-AI systems. \par
    
    Democratized learning provides unique ingredients to develop future distributed personalized intelligent systems. To that end, the learning design could be further studied with personalized datasets, extended for multi-task learning capabilities, and validated with actual generalization capabilities in practice for new users and environmental changes. 
We advocate our current design has not coped with multiple learning tasks \cite{mocha2017} and the adaptability of the system due to environmental changes as general intelligent systems. Turning the general distributed learning systems into reality, we need to profoundly analyze the Dem-AI from a variety of perspectives such as robustness and diversity of the learning models and novel knowledge transfer and distillation mechanisms \cite{TransferLearning2018,fed_distillation2018}. Also it is possible to incorporate our flexible design with current approaches such as meta-learning and optimization-based methods, to further improve the personalization in FL. We consider our work as an orthogonal contribution to the design architecture of distributed AIs with recent approaches. Besides, we believe it is necessary to experiment these design approaches in common personalized settings using realistic datasets for personalized learning tasks. 






\bibliographystyle{IEEEtran}
\bibliography{Dem-AI}

\ifCLASSOPTIONcaptionsoff
\fi


\begin{IEEEbiography}[{\includegraphics[width=1in,height=1.25in,clip,keepaspectratio]{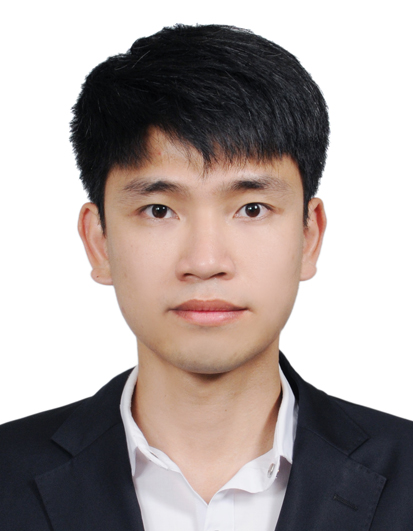}}]
	{\bf Minh N. H. Nguyen}  (M'20)
	received the BE degree in Computer Science and Engineering from Ho Chi Minh City University of Technology, Vietnam, in 2013 and the Ph.D. degree in Computer Science and Engineering from Kyung Hee University, Seoul, South Korea, in 2020. He is currently working as a lecturer at The University of Danang - Vietnam-
Korea University of Information and Communication Technology, Vietnam and a collaborator with the Networking Intelligence lab, Kyung Hee University, Korea. He received the best KHU PhD thesis award in engineering in 2020. His research interests include wireless communications, mobile edge computing, federated learning, and distributed machine learning.
\end{IEEEbiography}

\begin{IEEEbiography}[{\includegraphics[width=1in,height=1.25in,clip,keepaspectratio]{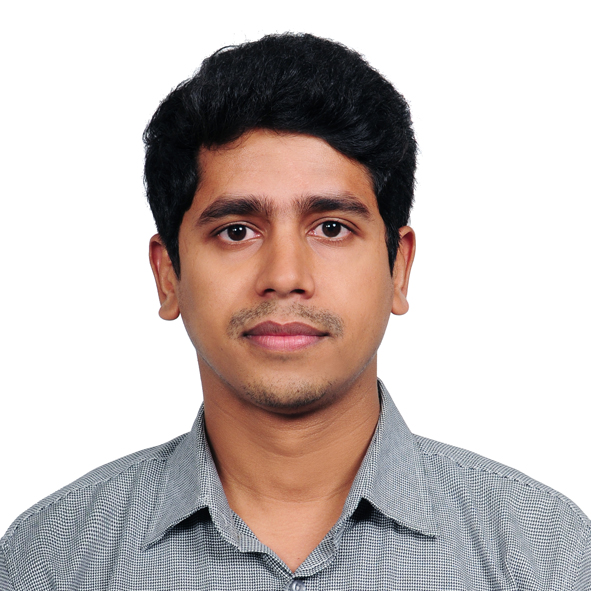}}]
	{\bf {Shashi Raj Pandey}}(M'21) is currently working as a Postdoctoral Researcher at the Connectivity Section, Aalborg University. He received his  B.E. degree in Electrical and Electronics with a specialization in Communication from Kathmandu University, Nepal in 2013, and the Ph.D. degree in Computer Science and Engineering from Kyung Hee University, Seoul, South Korea, in August 2021. He served as a Network Engineer at Huawei Technologies Nepal Co. Pvt. Ltd, Nepal from 2013 to 2016.  His research interests include network economics, game theory, wireless communications, data markets and distributed machine learning. 
\end{IEEEbiography}

\begin{IEEEbiography}[{\includegraphics[width=1in,height=1.25in,clip,keepaspectratio]{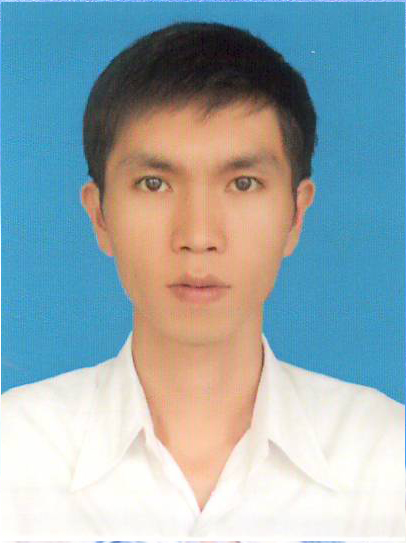}}]
	{\bf Tri Nguyen Dang} received the B.S. degree in information teaching from Hue University’s College of Education, Hue, Vietnam, in 2014. He is currently pursuing the Ph.D. degree in computer science and engineering with Kyung Hee University, Yongin,
    South Korea, under fully funded scholarship. His professional experiences include mobile application developer and middleware programming. His research interests include network optimization, mobile cloud computing, mobile-edge computing, and Internet of Things.
    Dr. Dang had won consolation prize of Olympiad in Informatics ACM ICPC Vietnam in 2013.
\end{IEEEbiography}

\begin{IEEEbiography}[{\includegraphics[width=1in,height=1.25in,clip,keepaspectratio]{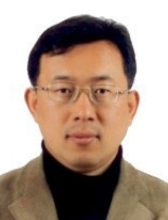}}]
	{\bf Eui-Nam Huh}
	(Member, IEEE) received the B.S. degree from Busan National University, South Korea, the master’s degree in computer science from The University of Texas, USA, in 1995, and the Ph.D. degree from Ohio University, USA, in 2002. He is currently a Professor with the Department of Computer Science and Engineering, Kyung Hee University, South Korea. His research interests include cloud computing, the Internet of Things, future Internet, distributed real-time systems, mobile computing, big data, and security. He is a member of the Review Board of the National Research Foundation of Korea. He has also served many community services for ICCSA, WPDRTS/IPDPS, APAN Sensor Network Group, ICUIMC, ICONI, APIC-IST, ICUFN, and SoICT as various types of chairs. He is also the Vice-Chairman of the Cloud/Bigdata Special Technical Group of TTA and an Editor of ITU-T SG13 Q17.
\end{IEEEbiography}

\begin{IEEEbiography}[{\includegraphics[width=1in,height=1.25in,clip,keepaspectratio]{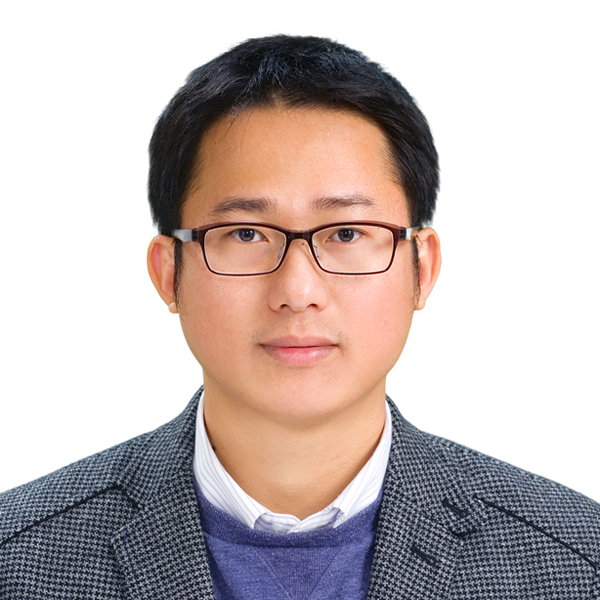}}]
	{\bf Nguyen H. Tran}
	(S\textquoteright10-M\textquoteright11-SM\textquoteright18)
	received BS and Ph.D degrees, from HCMC University of Technology and Kyung Hee University, in electrical and computer engineering, in 2005 and 2011, respectively. He was an Assistant Professor with Department of Computer Science and Engineering, Kyung Hee University, from 2012 to 2017. Since 2018, he has been with the School of Computer Science, The University of Sydney, where he is currently a Senior Lecturer. His research interests include distributed computing, machine learning, and networking. He received the best KHU thesis award in engineering in 2011 and several best paper awards, including IEEE ICC 2016 and ACM MSWiM 2019. He receives the Korea NRF Funding for Basic Science and Research 2016-2023 and ARC Discovery Project 2020-2023. He was the Editor of IEEE Transactions on Green Communications and Networking from 2016 to 2020, and the Associate Editor of IEEE Journal of Selected Areas in Communications 2020 in the area of distributed machine learning/Federated Learning.
\end{IEEEbiography}

\begin{IEEEbiography}[{\includegraphics[width=1in,height=1.25in,clip,keepaspectratio]{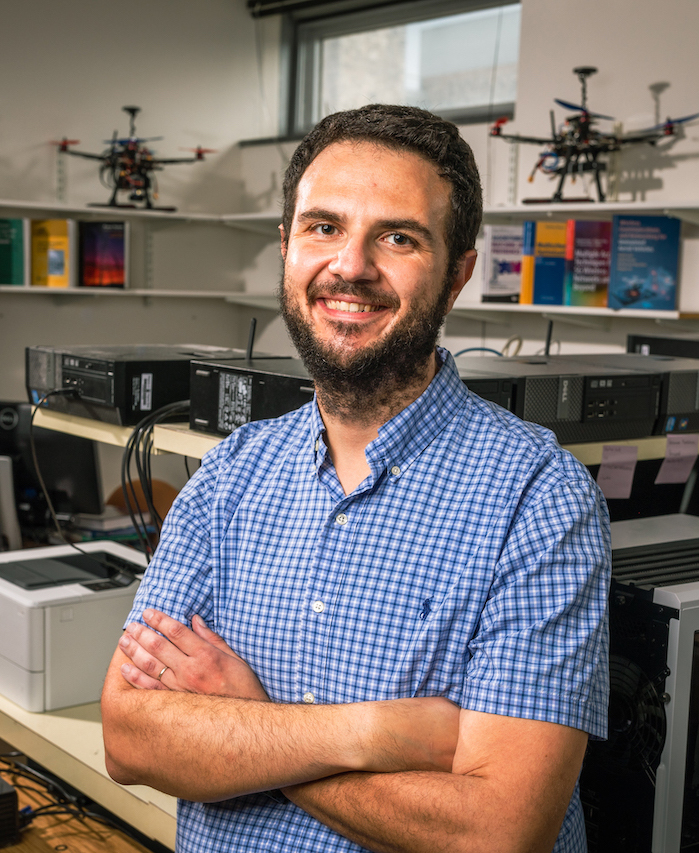}}]
	{\bf Walid Saad} (S\textquoteright07, M\textquoteright10, SM\textquoteright15, F\textquoteright19) received his Ph.D degree from the University of Oslo in 2010. Currently, he is a Professor at the Department of Electrical and Computer Engineering at Virginia Tech where he leads the Network sciEnce, Wireless, and Security (NEWS) laboratory. His research interests include wireless networks (5G/6G/beyond), machine learning, game theory, cybersecurity, unmanned aerial vehicles, semantic communications, and cyber-physical systems. Dr. Saad was the author/co-author of ten conference best paper awards and of the 2015 and 2022 IEEE ComSoc Fred W. Ellersick Prize. He was a co-author of the 2019 IEEE Communications Society Young Author Best Paper and of the 2021 IEEE Communications Society Young Author Best Paper. He is a Fellow of the IEEE. He currently serves as an editor for the IEEE Transactions on Mobile Computing and the IEEE Transactions on Cognitive Communications and Networking. He is an Area Editor for the IEEE Transactions on Network Science and Engineering, an Associate Editor-in-Chief for the IEEE Journal on Selected Areas in Communications (JSAC) – Special issue on Machine Learning for Communication Networks, and an Editor-at-Large for the IEEE Transactions on Communications.

\end{IEEEbiography}

\begin{IEEEbiography}[{\includegraphics[width=1in,height=1.25in,clip,keepaspectratio]{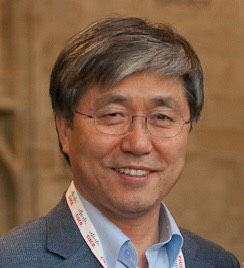}}]
	{\bf Choong Seon Hong}
	Choong Seon Hong (Senior Member, IEEE) received the B.S. and M.S. degrees in electronic engineering from Kyung Hee University, Seoul, South Korea, in 1983 and 1985, respectively, and the Ph.D. degree from Keio University, Japan, in 1997. In 1988, he joined KT, where he was involved in broadband networks as a Member of Technical Staff. He was with the Telecommunications Network Laboratory, KT, as a Senior Member of Technical Staff and as the Director of the Networking Research Team, until 1999. Since 1999, he has been a Professor at the Department of Computer Science and Engineering, Kyung Hee University. His research interests include machine learning and mobility management. He has served as the General Chair, the TPC Chair/Member, or an Organizing Committee Member of international conferences such as NOMS, IM, APNOMS, E2EMON, CCNC, ADSN, ICPP, DIM, WISA, BcN, TINA, SAINT, and ICOIN. He was an Associate Editor of the IEEE Transactions on Network and Service Management and Journal Communications and Networks, and an Associate Technical Editor of the IEEE Communications Magazine. He currently serves as an Associate Editor of the International Journal of Network Management and Future Internet. He is a member of ACM, IEICE, IPSJ, KIISE, KICS, KIPS, and OSIA.			
\end{IEEEbiography}

\end{document}